\documentclass[journal,letterpaper]{IEEEtran}  
\usepackage{amsmath,amsfonts}
\usepackage{algorithmic}
\usepackage{algorithm}
\usepackage{array}
\usepackage[caption=false,font=normalsize,labelfont=sf,textfont=sf]{subfig}
\usepackage{textcomp}
\usepackage{stfloats}
\usepackage{url}
\usepackage{verbatim}
\usepackage{multirow}
\usepackage{graphicx}
\usepackage{float}
\usepackage{booktabs}
\usepackage{cite}
\usepackage{xcolor}
\usepackage{balance}
\usepackage{amssymb}
\usepackage{pifont}
\usepackage{tabularx}
\newcolumntype{Y}{>{\centering\arraybackslash}X}

\def\BibTeX{{\rm B\kern-.05em{\sc i\kern-.025em b}\kern-.08em
    T\kern-.1667em\lower.7ex\hbox{E}\kern-.125emX}}

\begin{document}
\title{\textcolor{black}{ZTFed-MAS2S: A Zero-Trust Federated Learning Framework with Verifiable Privacy and Trust-Aware Aggregation for Wind Power Data Imputation}}
\author{Yang Li, \textit{Senior Member, IEEE}, \and Hanjie Wang, \and Yuanzheng Li,
 \textit{Senior Member, IEEE}, \and \\Jiazheng Li, \and Zhaoyang Dong, \textit{Fellow, IEEE}
 \thanks{Y. Li and H. Wang are with the School of Electrical Engineering, Northeast Electric Power University, Jilin 132012, China (e-mail: liyang@neepu.edu.cn; wanghanjie1@163.com).
 \par Y. Z. Li is with the School of Artificial Intelligence and Automation, Huazhong University of Science and Technology, Wuhan 430074, China (email: Yuanzheng\_Li@hust.edu.cn).
\par J. Z. Li is with the State Grid Xiamen Electric Power Supply Company, Xiamen 361004, China (email: tytrytytgr@gmail.com).
\par Z. Y. Dong is with the Department of Electrical Engineering, City University of Hong Kong, Kowloon, Hong Kong, (email: zydong@cityu.edu.hk).

 }}

\maketitle
\textcolor{black}{\begin{abstract}
Wind power data often suffers from missing values due to sensor faults and unstable transmission at edge sites. While federated learning enables privacy-preserving collaboration without sharing raw data, it remains vulnerable to anomalous updates and privacy leakage during parameter exchange. These challenges are amplified in open industrial environments, necessitating zero-trust mechanisms where no participant is inherently trusted. To address these challenges, this work proposes ZTFed-MAS2S, a zero-trust federated learning framework that integrates a multi-head attention-based sequence-to-sequence imputation model. ZTFed integrates verifiable differential privacy with non-interactive zero-knowledge proofs and a confidentiality and integrity verification mechanism to ensure verifiable privacy preservation and secure model parameters transmission. A dynamic trust-aware aggregation mechanism is employed, where trust is propagated over similarity graphs to enhance robustness, and communication overhead is reduced via sparsity- and quantization-based compression. MAS2S captures long-term dependencies in wind power data for accurate imputation. Extensive experiments on real-world wind farm datasets validate the superiority of ZTFed-MAS2S in both federated learning performance and missing data imputation, demonstrating its effectiveness as a secure and efficient solution for practical applications in the energy sector.
\end{abstract}
}


\begin{table}[!t]
\centering
\footnotesize\textbf{NOMENCLATURE}\par\vspace{4pt} % 手写标题，无编号
\renewcommand{\arraystretch}{1.05}
\begin{tabularx}{\linewidth}{@{}p{2.4cm}X@{}}
ZTFed   & zero-trust federated learning framework \\
-MAS2S  & with a multi-head attention-based sequence-to-sequence imputation model \\
FL      & federated learning \\
DP      & differential privacy \\
MA      & multi-head attention \\
NIZK    & non-interactive zero-knowledge proofs \\
CIV     & Confidentiality and Integrity Verification \\
DTAA    & Dynamic Trust-Aware Aggregation \\
Bi-LSTM & bidirectional long short-term memory \\
IIoT    & industrial internet of things scenarios \\
GAIN    & generative adversarial imputation nets \\
FHE     & fully homomorphic encryption \\
TSS     & threshold secret sharing \\
T-Mean  & Trimmed Mean \\
MAAPE   & arctangent absolute percentage error \\
NREL    & National Renewable Energy Laboratory \\
AES-CBC & Advanced Encryption Standard in Cipher Block Chaining mode \\
HMAC    & Hash-based Message Authentication Code \\
MAD     & median absolute deviation \\
MIA     & membership inference attacks \\
\end{tabularx}
\end{table}

\section{Introduction}
\IEEEPARstart{T}{oday}, as global energy demand grows and the emphasis on sustainable environmental development intensifies, the advancement and integration of renewable energy, particularly wind power, have become key trends. Wind power generation, the most mature and broadly commercialized form of renewable energy, significantly impacts the power system as it advances and integrates into the grid, due to its randomness, volatility, and intermittency~\cite{aaslid2022stochastic}. Historical operational data from wind farms is crucial for studying wind power dynamics, improving forecasting accuracy, assessing its impact on the grid, and developing effective control strategies~\cite{10419905, 10839636, 10603406}. However, during the data collection, transmission, and conversion processes, especially at network edge wind farms, frequent data missing occurs due to sensor failures, transmission problems, and noise interference. This degrades measurement quality, complicates data analysis, and impacts power grid operations. Furthermore, how to achieve secure and efficient data sharing while ensuring privacy for wind power missing data imputation is increasingly becoming a crucial and urgent topic~\cite{9609984}.

\vspace{-0.3cm}
\subsection{Literature Review}  
Based on previous research on wind data imputation, existing methods can be categorized into three types: statistical, traditional machine learning, and deep learning. Statistical methods include value interpolation and modeling. Mean imputation offers a simple implementation. Modeling approaches such as autoregressive (AR)~\cite{9548664} and expectation-maximization (EM)~\cite{10106036} assume specific data distributions for imputation. Traditional machine learning methods, including k-nearest neighbors (kNN)~\cite{10460131} and Missforest~\cite{9518376}, build predictive models based on existing features and are effective in handling nonlinear relationships and large-scale data.

While traditional methods have made progress, deep learning techniques have demonstrated even greater potential. For example, reference~\cite{liu2022missing} proposes a super-resolution perception convolutional neural network that integrates super-resolution techniques to reconstruct missing data from low-resolution inputs; reference~\cite{10843972} introduces a multiscale spatiotemporal Transformer model for industrial process imputation; reference~\cite{10534859} develops an attention-based network leveraging polynomial regression for solar photovoltaic power data; reference~\cite{10144489} proposes an adaptive multi-head self-attention-based variational auto-encoder model for imputation; reference~\cite{10636213} designs an unsupervised conditional generative adversarial network model using data-mining techniques for energy data imputation; and reference~\cite{li2022integrated} puts forward an integrated long short-term memory (LSTM) model for missing data tolerance.

Despite progress in missing data imputation, challenges remain in capturing complex dependencies, long-sequence features, and dynamic variation patterns. Traditional CNN- and RNN-based models struggle with long-range dependencies and generalization, particularly on large-scale, high-dimensional data. In parallel, centralized learning methods, though benefiting from aggregated data, face practical issues such as privacy risks in cross-farm data integration and high communication overhead between edge nodes and central servers.

Federated learning (FL) enables collaborative model training by sharing parameters instead of raw data, mitigating privacy and communication challenges compared to centralized approaches~\cite{li2023wind}. Consequently, it has been widely adopted in industrial Internet of Things (IIoT) scenarios~\cite{10579800, li2022detection, fotohi2024decentralized}. \textcolor{black}{However, existing FL methods still face notable limitations. First, model parameters may exhibit anomalies during aggregation, such as integrity violations from network issues or adversarial manipulations like backdoor attacks, compromising system performance and security. To address this, robust aggregation mechanisms such as MultiKrum~\cite{10121613}, Trimmed Mean (T-mean)~\cite{wang2025federated}, and Median~\cite{9721118} have been employed. Yet, these rely on fixed thresholds (e.g., adversary ratios, trimming rates), often excluding benign clients or failing to detect sophisticated anomalies. Second, despite avoiding raw data sharing, FL still faces privacy risks during parameter transmission. Techniques such as differential privacy (DP), fully homomorphic encryption (FHE)~\cite{hijazi2023secure}, and threshold secret sharing (TSS)~\cite{8765347} attempt to address these risks via noise injection or encryption, but rely on the assumption of honest client behavior—often unrealistic in practice. In open and decentralized settings, these challenges highlight the need for zero-trust (ZT) mechanisms, where no entity is inherently trusted~\cite{stafford2020zero}.}

In summary, there are three key gaps in wind power missing data imputation:

1) \textcolor{black}{A distributed learning approach ensuring privacy preservation and verifiability is needed under a ZT architecture to address the privacy risks and high communication overhead associated with edge-based wind farms.}

2) The lack of advanced models capable of capturing long-time dependencies and dynamic changes in time series data for missing data imputation, particularly for wind power data.

3) Existing methods for constructing missing data scenarios often consider discrete or continuous missing patterns individually, but seldom address the hybrid of both simultaneously. This leads to difficulties in capturing the complexity of the coexistence of patterns in real-world scenarios.

\subsection{Contribution of This Paper} 

This paper proposes ZTFed-MAS2S, a zero-trust federated learning framework with a multi-head attention-based sequence-to-sequence imputation model for wind power data. The key contributions are as follows:

1) \textcolor{black}{The ZTFed framework integrates verifiable Differential Privacy with Non-Interactive Zero-Knowledge Proofs (DP-NIZK) and a Confidentiality and Integrity Verification (CIV) mechanism to enable verifiable privacy preservation and secure, integrity-assured model transmission. In addition, it employs a Dynamic Trust-Aware Aggregation (DTAA) mechanism to enhance resilience against anomalous clients and incorporates sparsity- and quantization-based compression to reduce communication overhead. }

2) \textcolor{black}{The MAS2S imputation model leverages wind features as inputs to provide richer contextual information.} It incorporates a bidirectional long short-term memory (Bi-LSTM) encoder and an LSTM decoder, \textcolor{black}{enhanced by a multi-head attention (MA) mechanism. The MA mechanism in the decoder effectively captures the correlations between wind power and other features, thereby enhancing the accuracy of the final output.}

3) To simulate hybrid missing patterns in missing data scenarios, we propose a hybrid masking strategy. By designing a missing matrix, this strategy precisely coordinates discrete missing points and continuous missing intervals, enabling more realistic simulation of complex real-world scenarios.

4) \textcolor{black}{Extensive experiments on real-world wind farm datasets were conducted, including comparisons with state-of-the-art privacy-preserving federated learning frameworks—covering both privacy and aggregation mechanisms—as well as with traditional imputation methods. To the best of our knowledge, this is the first FL-based imputation framework to integrate verifiable differential privacy and dynamic trust-aware aggregation under a zero-trust architecture.}

\section{Methods}

\subsection{\textcolor{black}{MAS2S Imputation Model}}
The key feature of the sequence-to-sequence model is its end-to-end learning \cite{bahdanau2014neural}, which directly maps input sequences to output sequences without relying on observation windows or prediction windows. This makes it ideal for multivariate time series analysis \cite{10111057}. {Fig. 1 illustrates its structure. 

\begin{figure}[ht]
    \centering
    \includegraphics[width=1\linewidth]{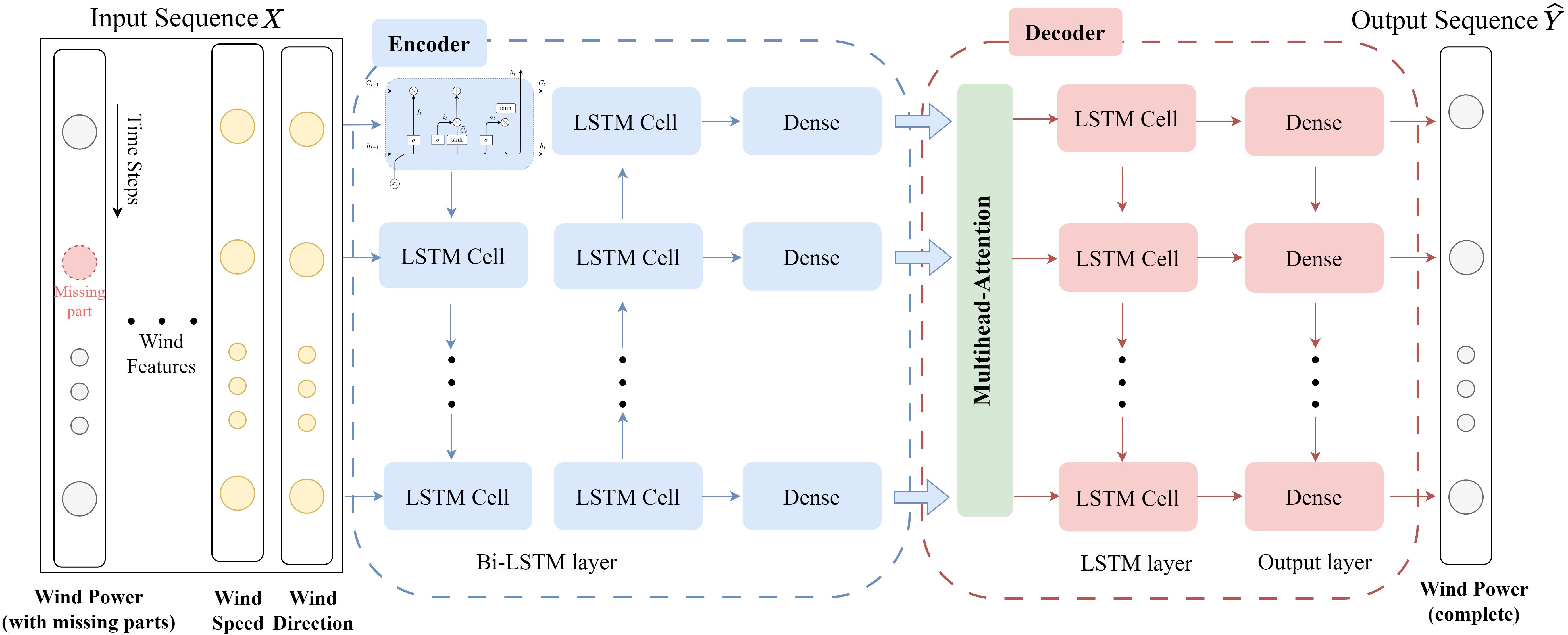}
    \caption{{The basic structure of the \textcolor{black}{MAS2S}.}}
    \label{fig:enter-label1}
%    \vspace{-0.5cm}
\end{figure}

As shown in the figure, the input sequence $X$ and the output sequence $\widehat {Y}$ are represented as follows:}
\begin{equation}
    X = \{x_1, \ldots, x_t,\ldots,x_T\},\quad 
    \widehat {Y} = \{\widehat {y_1}, \ldots, \widehat {y_t}, \ldots, \widehat {y_T}\}
\end{equation}
where $T$ represents the total number of time steps, $x_t$ denotes the input feature vector at time step $t$, which includes the wind power data with missing parts and other wind features (e.g., wind speed and wind direction), $\widehat {y}$ represents the output wind power data, with missing parts imputed.

The MAS2S adopts an encoder-decoder architecture, where a Bi-LSTM encoder maps the input time series to a semantic vector capturing global context, and a decoder with MA and LSTM generates the target sequence based on feature correlations.

\subsubsection{BiLSTM Encoder}
In each $t$, the LSTM maintains its internal hidden state $h$, which results in a hidden sequence of $\{h_1, h_2,\ldots, h_T\}$. The hidden vector $h_t$ is updated as follows:
\begin{equation}
    {f_t} = sigmoid \left( {{W_f} \left[ {{h_{t - 1}}\parallel{x_t}} \right] + {b_f}} \right),
    \label{eq:ft}
\end{equation}
\begin{equation}
    {i_t} = sigmoid \left( {{W_i} \left[ {{h_{t - 1}}\parallel{x_t}} \right] + {b_i}} \right),
    \label{eq:it}
\end{equation}
\begin{equation}
{o_t} = sigmoid \left( {{W_o} \left[ {{h_{t - 1}}\parallel{x_t}} \right] + {b_o}} \right),
\end{equation}
\begin{equation}
\widetilde {{C_t}} = \tanh \left( {{W_C}\left[ {{h_{t - 1}}\parallel{x_t}} \right] + {b_C}} \right),
\end{equation}
\begin{equation}
{C_t} = {f_t}\;*{C_{t - 1}} + {i_t}*\widetilde {{C_t}},\quad {h_t} = {o_t}*\tanh \left( {{C_t}} \right).
\end{equation}
Here, ${f_t}$, ${i_t}$, and ${o_t}$ represent the forget gate, input gate, and output gate, respectively; $W_f$, $W_i$, and $W_o$ denote the \textcolor{black}{weight} matrices; $b_f$, $b_i$, and $b_o$ are the bias vectors; $\widetilde {{C_t}}$ and ${C_t}$ represent the cell vector and the candidate cell vector, respectively; ${h_t}$ refers to the encoder hidden state; the symbol $\|$ denotes concatenation.

In a BiLSTM, rather than generating a single sequence of hidden states, it computes both forward and backward hidden states, denoted by $\overrightarrow{h_t}$ and $\overleftarrow{h_t}$, respectively.
These are subsequently concatenated to form the hidden state:
\begin{equation}
    {h_t} =[ \mathop {{h_t}}\limits^ \to\parallel \mathop {\;{h_t}}\limits^ \leftarrow ].
\end{equation}
\subsubsection{\textcolor{black}{LSTM Decoder with Multi-Head Attention}}
\textcolor{black}{To enhance the decoder’s ability to capture temporal dependencies, we employ a MA mechanism. The decoder’s previous hidden state $s_{t-1}$ and the encoder hidden states $\{ h_1, \dots, h_T \}$ are projected into $N_h$ parallel subspaces.}

\textcolor{black}{For each attention head $m \in \{1, \dots, N_h\}$, we compute:}
\begin{equation}
\textcolor{black}{Q^m = W_Q^m s_{t-1}, \quad K^m = W_K^m h_i, \quad V^m = W_V^m h_i,}
\end{equation}
\textcolor{black}{where $Q^m$, $K^m$, and $V^m$ are the query, key, and value vectors, and $W_Q^m$, $W_K^m$, and $W_V^m$ are the corresponding projection matrices.}

\textcolor{black}{The attention weights $\alpha_{t,i}^m$ are computed via scaled dot-product attention:}
\begin{equation}
\textcolor{black}{\alpha_{t,i}^m = \text{softmax} \left( \frac{Q^m (K^m)^\top}{\sqrt{d_k}} \right),}
\end{equation}
\textcolor{black}{where $d_k$ is the key dimension. The head-specific context vector $c_t^{a,m}$ is obtained as a weighted sum of the value vectors:}
\begin{equation}
\textcolor{black}{c_t^{a,m} = \sum_{i=1}^{T} \alpha_{t,i}^m V^m.}
\end{equation}

\textcolor{black}{Finally, the outputs of all heads are concatenated and linearly projected to form the final context vector $c_t^a$:}
\begin{equation}
\textcolor{black}{c_t^a = W_O \cdot [c_t^{a,1} \parallel \cdots \parallel c_t^{a,N_h} ].}
\end{equation}
\textcolor{black}{where $W_O$ is a projection matrix.} The decoder updates its hidden state $s_t$ and output $\hat{y}_t$ as:
\begin{equation}
s_t = \text{LSTM}(c_t^a\parallel s_{t-1}), \quad
\hat{y}_t = \text{Dense}(W_d\cdot{s_t} + b_d).
\end{equation}

To maintain the consistency and integrity of the data, we considered both the missing and non-missing parts of the output sequence for correction, defining the loss function for the entire sequence:
\begin{equation}
\mathcal{L}({\theta}) = \frac{1}{T}\sum\limits_{t = 1}^T {\left| {{y_t} - \widehat {{y_t}}} \right|} ,
\end{equation}
where \( \mathcal{L}(\cdot) \) is the loss function, ${y_t}$ represents the ground truth at time step $t$, and $\theta$ represents the model parameters.

The model update process can be expressed as:
\begin{equation}
{\theta} \leftarrow {\theta} - \eta  \cdot Adam({\theta},\nabla \mathcal{L}({\theta}),{\beta _1},{\beta _2}),
\end{equation}
where \( \eta \) is the learning rate; \( {\beta_1} \) and \( {\beta_2} \) are the momentum parameters; \( \theta \) represents the MAS2S imputation model parameters; and \( \nabla \mathcal{L}({\theta}) \) denotes the gradient of the loss function.

The overall algorithmic flow of the MAS2S imputation model is shown in Algorithm 1.

\begin{algorithm}[!ht] % ！htbp
    \small
    \caption{MAS2S imputation model} %算法标题
    \label{alg:mas2s}
    \begin{algorithmic}[1] %[] 里面的数字为行号 [1]就是一行一个数字
    \REQUIRE Training epoch $I_l$ and its index $i_l$; input sequence $X$; ground truth sequence $Y$.
    \ENSURE MAS2S imputation model with initialized parameters $\theta$.
    \FOR{epoch $i_l=1,2,\ldots, I_l$} 
    \FOR{batch of input sequences $X$} 
    \STATE Run encoding by feeding sequence $X$ into Bi-LSTM according to (2)-(7).
    \STATE Obtain the entire hidden state sequence $\{ {h_1},{h_2}, \ldots ,{h_T}\} $.
    \STATE Run decoding by computing attention vector $ c_t^a $ according to (8)-(11).
    \STATE Obtain the output sequence $\widehat{Y}$ according to (12).
    \STATE Calculate the loss according to (13).
    \STATE Update parameters according to (14).
    \ENDFOR
    \ENDFOR
    
\end{algorithmic}
\end{algorithm}

\subsection{\textcolor{black}{Zero-Trust Federated Learning Framework}}
\subsubsection{\textcolor{black}{Differential Privacy with Non-Interactive Zero-Knowledge Proofs}}
\textcolor{black}{To achieve both privacy preservation and verifiability in FL, we propose DP-NIZK. DP protects client data by adding noise to model parameters. NIZK enables the server to verify the correct application of DP noise without disclosing sensitive information~\cite{10966041}.}

\textcolor{black}{DP guarantees that for any adjacent datasets $\mathcal{D}_i$ and $\mathcal{D}_i'$, and for any measurable set $\mathcal{S}$ in the output space, the randomized mechanism $\mathcal{M}$ satisfies:}
\begin{equation}
\textcolor{black}{\Pr[\mathcal{M}(\mathcal{D}_i) \in \mathcal{S}] \leq e^\epsilon \Pr[\mathcal{M}(\mathcal{D}_i') \in \mathcal{S}] + \delta,}
\end{equation}
where $\delta$ represents the probability of privacy leakage, $\epsilon$ is the privacy budget, \textcolor{black}{and $\Pr[\cdot]$ denotes the probability over the randomness of $\mathcal{M}$. To achieve global $(\epsilon,\delta)$-DP, we employ the Gaussian mechanism~\cite{9069945}.} The local model parameters $\theta_i$ are first clipped to bound their $\ell_2$-norm:
\begin{equation}
\bar{\theta}_i = \theta_i / \max(1, \|\theta_i\|_2/\tau_c),
\end{equation}
where \( \bar{\theta}_i \) denotes the clipped parameters of client $i$, and $\tau_c$ is the clipping threshold. Gaussian noise \( \mathbf{n}_i \) is then added:
\begin{equation}
\widetilde{\theta}_i = {\bar \theta _i} + {\bf{n}}_i\sim \mathcal{N}(0, \sigma^2).
\end{equation}
\textcolor{black}{Here, ${\widetilde {{\theta}_i}}$ denotes the perturbed model parameters, and the standard deviation $\sigma$ of the Gaussian noise is determined by:}
\begin{equation}
\textcolor{black}{\sigma = \frac{\sqrt{2\ln(1.25/\delta)} \cdot \Delta s}{\epsilon}\times \frac{T_g}{K}.}
\end{equation}
\textcolor{black}{Here, $T_g$ denotes the global epoch, and $K$ is the synchronization interval. The global sensitivity $\Delta s$ is defined as the maximum local sensitivity across all clients:}
\begin{equation}
\textcolor{black}{\Delta s = \max \left\{ \Delta s_i \right\}, \quad \forall i,}
\end{equation}
\textcolor{black}{where $\Delta s_i$ denotes the local sensitivity of client $i$, computed as the maximum change in the local model output when the dataset changes from $\mathcal{D}_i$ to $\mathcal{D}_i'$, with the local model output defined as ${s^{{{\cal D}_i}}} = \mathop {\arg \min }\limits_{{\bar{\theta} _i}} {\cal L}({\bar{\theta} _i},{{\cal D}_i})$. The local sensitivity is:
}
\begin{equation}
\textcolor{black}{\Delta s_i = \max_{\mathcal{D}_i,\mathcal{D}_i'} \left\| s^{\mathcal{D}_i} - s^{\mathcal{D}_i'} \right\|_2 = 2\tau_c / |\mathcal{D}_i|,}
\end{equation}

\textcolor{black}{After determining the $\sigma$, the client deterministically generates Gaussian noise from a private seed \( s \in \mathbb{Z}_q \), which serves as the secret witness. To construct the NIZK, the client performs three steps—commitment, challenge, and response—starting with computing commitments using \( s \) and a random nonce \( k \in \mathbb{Z}_q \):}
\begin{equation}
\textcolor{black}{h_s = g^{s} \bmod p, \quad t_k = g^{k} \bmod p.}
\end{equation}
\textcolor{black}{Here, \( h_s \) and \( t_k \) denote the public and temporary commitments, respectively. The parameters \( p \) and \( q \) are large primes, and \( g \in \mathbb{Z}_p^* \) is a generator of a subgroup of order \( q \).}

\textcolor{black}{Next, the challenge \( c_s \) is autonomously derived by hashing the commitments and parameters using a collision-resistant hash function \( H(\cdot) \):}
\begin{equation}
\textcolor{black}{c_s = H(t_k \parallel H(\widetilde{\theta}_i) \parallel H(\theta_i)) \bmod q.}
\end{equation}

\textcolor{black}{With the challenge determined, the client computes the corresponding response $r_s$:}
\begin{equation}
\textcolor{black}{r_s = (k + c_s \cdot s) \bmod q.}
\end{equation}

\textcolor{black}{Finally, the client uploads the NIZK proof \( p_f=\{t_k, r_s, c_s, h_s, H(\widetilde{\theta}_i), H(\theta_i)\} \) along with \( \widetilde{\theta}_i \). Upon reception, the server verifies whether:}
\begin{equation}
    \textcolor{black}{g^{r_s} \bmod p =t_k \cdot h^{c_s} \bmod p.}
\end{equation}

\textcolor{black}{If verification succeeds, it confirms correct noise application, and the server accepts \( \widetilde{\theta}_i \) as valid model parameters for aggregation. Otherwise, the parameters are deemed untrustworthy and excluded.}

\subsubsection{\textcolor{black}{Communication-Efficient and Secure Transmission}}  

\textcolor{black}{To reduce communication overhead while ensuring privacy-preserving and verifiable model parameter transmission, we propose a framework that integrates sparsity- and quantization-based compression with a Confidentiality and Integrity Verification (CIV) mechanism.}

\textcolor{black}{The transmitted parameters $\theta^c \in \{\widetilde{\theta}, \theta_g\}$, where \( \theta_g \) denotes the aggregated global model parameters, are compressed as:}
\begin{equation}
\textcolor{black}{\theta^c = Q_b\left( S_p(\theta^c) \odot M^{\text{com}} \right),}
\end{equation}
\textcolor{black}{where \( \theta^c \) denotes the compressed parameters; \( S_p(\cdot) \) is a sparsification function that selects the top-\( p\% \) entries with the largest magnitudes; \( Q_b(\cdot) \) quantizes the selected values into \( b \)-bit integers; \( M^{\text{com}} \) is a binary mask indicating the retained positions; \( \odot \) denotes the Hadamard product.}

\textcolor{black}{Then, \( \theta^c \) is encrypted using a symmetric key \( k_{\text{sym}} \), which is assumed to be securely established under a zero-trust architecture. To ensure integrity, a Hash-based Message Authentication Code (HMAC) is appended:}
\begin{equation}
\textcolor{black}{\mathcal{M}_{\text{enc}} = \text{AES-CBC}(k_{\text{sym}}, \theta^c) \, \| \, \text{HMAC}(\theta^c),}
\end{equation}
\textcolor{black}{where AES-CBC stands for the Advanced Encryption Standard in Cipher Block Chaining mode, and $\mathcal{M}_{enc} \in \{\mathcal{M}_{enc}^U, \mathcal{M}_{enc}^D\}$ denotes the encrypted transmission message, where $\mathcal{M}_{enc}^U$ and $\mathcal{M}_{enc}^D$ refer to the client’s encrypted upload and download messages, respectively.}

\textcolor{black}{Upon receiving \( \mathcal{M}_{\text{enc}} \), decryption with \( k_{\text{sym}} \) yields \( \theta^c \), and integrity is verified by recomputing the HMAC. If successful, the original parameter is reconstructed as:}
\begin{equation}
\textcolor{black}{\theta = Q_b^{-1}(\theta^c).}
\end{equation}

\subsubsection{\textcolor{black}{Dynamic Trust-Aware Aggregation}}

\textcolor{black}{We propose DTAA to enable robust aggregation under a ZT architecture. DTAA estimates inter-client trust through similarity analysis and graph-based propagation in each aggregation round, and then selects clients for aggregation based on trust scores.}

\textcolor{black}{First, the cosine similarity between client model parameters is computed as:}
\begin{equation}
\textcolor{black}{S_{i,j} = {\rm{cos}}({\tilde \theta _i} , {\tilde \theta _j}) =\frac{\tilde{\theta}_i \cdot \tilde{\theta}_j}{\|\tilde{\theta}_i\|_2 \cdot \|\tilde{\theta}_j\|_2}, \quad i \neq j,}
\end{equation}
\textcolor{black}{where $S_{i,j}$ indicates the similarity between clients $i$ and $j$. To enhance contrast, we apply a power transformation $T^s_{i,j} = (S_{i,j})^r$ where $r > 0$ is a sharpening coefficient, and $T^s_{i,j} \in [0,1]$ is the normalized trust score.}

\textcolor{black}{Then, a sparse trust graph \( G' = (V, E', T^s ) \) is constructed, where \( V \) is the client set, and \( E' \subseteq V \times k_g \) contains directed edges from each client \( i \in V \) to its top-\( k_g \) most trusted neighbors based on \( T_{i,j}^s \). This graph defines a weighted adjacency matrix \( A \in \mathbb{R}^{N \times N} \) as:}
\begin{equation}
\textcolor{black}{A_{i,j} =         
\begin{cases}
T_{i,j}, & \text{if } (i \rightarrow j) \in E', \\
0, & \text{otherwise}.                                                    
\end{cases}          }    
\end{equation}

\textcolor{black}{The initial trust score of client \( i \) is then computed as:}
\begin{equation}
\textcolor{black}{\mathbf{t}_i^{(0)} = \frac{\sum_j A_{i,j}}{\sum_{i,j} A_{i,j}}.}
\end{equation}

\textcolor{black}{Trust scores are iteratively updated via propagation with damping:}
\begin{equation}
\textcolor{black}{\mathbf{t}_i^{(t+1)} = (1 - d) A    \mathbf{t}_i^{(t)} + d \mathbf{t}_i^{(0)},}
\end{equation}
\textcolor{black}{where \( d \in (0,1) \) is the damping factor. The process converges when:}
\begin{equation}
\textcolor{black}{\left| \mathbf{t}^{(t+1)} - \mathbf{t}^{(t)} \right| < \tau_t,}
\end{equation}
\textcolor{black}{where \( \tau_t > 0 \) is a convergence threshold, and \( \mathbf{t}^{(t)} = [\mathbf{t}_1^{(t)}, \dots, \mathbf{t}_N^{(t)}]^\top \in \mathbb{R}^N \) denotes the trust scores of all clients at iteration \( t \).}

\textcolor{black}{To enhance robustness, an anomaly detection step based on the median absolute deviation (MAD) is introduced:}
\begin{equation}
\textcolor{black}{\text{MAD} = \text{Median}(|\mathbf{t}_i - \text{Median}(\mathbf{t})|) ,}
\end{equation}

\textcolor{black}{Based on this, clients with \(\mathbf{t}_i <  \text{Median}(\mathbf{t}) - k_m \cdot \text{MAD}\) are excluded, where \( k_m \) is a tuning coefficient. }

\textcolor{black}{Finally, the aggregation is performed in a layer-wise manner, applying a median operation over the selected client set \(V_{\text{s}}\):}
\begin{equation}
\textcolor{black}{\theta_g = \text{Median}(\{\widetilde{\theta}_i \mid i \in V_{\text{s}}\}).}
\end{equation}

\section{ZTFed-MAS2S For Wind Power Imputation}

\subsection{Problem Formulation}
{For the input sequence \(X\), it can be partitioned into an observed part \(X_{\text{obs}}\) containing all the observed data components, and a missing part \(X_{\text{miss}}\) that includes all the missing data, i.e., \(X = \{X_{\text{obs}}, X_{\text{miss}}\}\). The objective is to derive a complete sequence \(\widehat{Y}\) where the missing entries are imputed using the observed data. This task can be formalized as follows:}
\begin{equation}
\widehat {{Y}} = impute(X_{\text{miss}} | X_{\text{obs}}),
\end{equation}

\noindent where, $impute( \cdot )$ represents an imputation method.

\subsection{{Simulation of Missing Scenarios}}
The construction of missing scenarios is driven by two factors: missing rate (proportion of absent values) and missing patterns (continuous or discrete), where discrete missing refers to missing completely at random. To emulate real-world challenges in wind farm data, our method combines these factors using a hybrid masking strategy to obtain the deteriorated data. As illustrated in Fig. 2, this process includes three key steps:

First, a total missing rate \(mr_t\) is defined for the complete wind power sequence \(X^w = \{x_1^w, \dots, x_t^w, \dots, x_T^w\}\). A tunable ratio parameter \(r_m \in [0,1]\) is then introduced to partition \(mr_t\) into continuous and discrete components. Specifically:
{\begin{equation}
\left\{ \begin{array}{l}
mr_d = r_m \cdot mr_t, \\
mr_c = (1 - r_m) \cdot mr_t,
\end{array} \right. 
\end{equation}}
where $mr_d$ and $mr_c$ represent the discrete and continuous missing rates, respectively.

Next, the continuous missing mask matrix \(M_c\) and the discrete missing mask matrix \(M_d\) are generated using random sampling based on the missing rates $mr_c$ and $mr_d$, and then they are concatenated as a hybrid mask matrix \(M_h\). In these matrices, 0 indicates complete data; 1 indicates missing data, with green and yellow denoting continuous and discrete patterns.

Finally, missing positions are identified by \(M_h\), missing values are injected into \(X_w\) to generate the deteriorated incomplete wind power sequence \(\widetilde {X_w}\):
\begin{equation}
\widetilde {X^w} = X^w \odot (1 - M_h) + pValue \cdot M_h,
\end{equation}
where \(pValue\) is a placeholder value, ensuring true zeros (e.g., no wind power generation) are not confused with missing data.

\subsection{Data Preprocessing}

To eliminate scale differences between variables and fairly account for the influence of each wind feature on wind power, all features are normalized using min-max normalization as:
\begin{equation}
x^{j'}_t=\frac{x_t^j-\min (x^j)}{\max (x^j)-\min (x^j)},
\end{equation}
where \( x_t^j \) is the original data of the \(j\)-th feature at time step $t$, and \( x^{j'}_t \) is the normalized value.

% 2) Outlier Handling: {This study uses an improved isolation forest (iForest) algorithm, which is well-suited for large-scale wind power datasets due to its ability to identify sparse outliers with minimal parameter tuning \cite{10711296}. The process starts by removing outliers that exceed wind power constraints using a physical threshold filter. This step helps minimize false anomaly detections, which may occur when data points lie near the boundary.}

% {Then, the iForest constructs multiple decision trees to isolate data points and calculates an anomaly score based on the path length:}
% {\begin{equation}
%     s(x_t) = 2^{ - \frac{h(x_t)}{c(z)}} 
% \end{equation}}
% {where $s(\cdot)$ represent the anomaly score function. \( h(x_t) \) is the average path length of \( x_t \) in the trees, and \( c(z) \) represents the expected path length to a leaf node containing $z$ data points. Then, $c(z)$ is calculated by}
% {\begin{equation}
% c(z) = 2 \left( \ln(z-1) + \gamma \right)
% \end{equation}}
% {where \( z \) is the number of data points and \( \gamma \) represents the Euler-Mascheroni constant. Anomalous points have shorter path lengths, resulting in higher anomaly scores \( s(x_t) \), and are classified as outliers. These outliers are flagged as missing values for subsequent imputation.}

\begin{figure}[!t]
    \centering
    \includegraphics[width=1\linewidth]{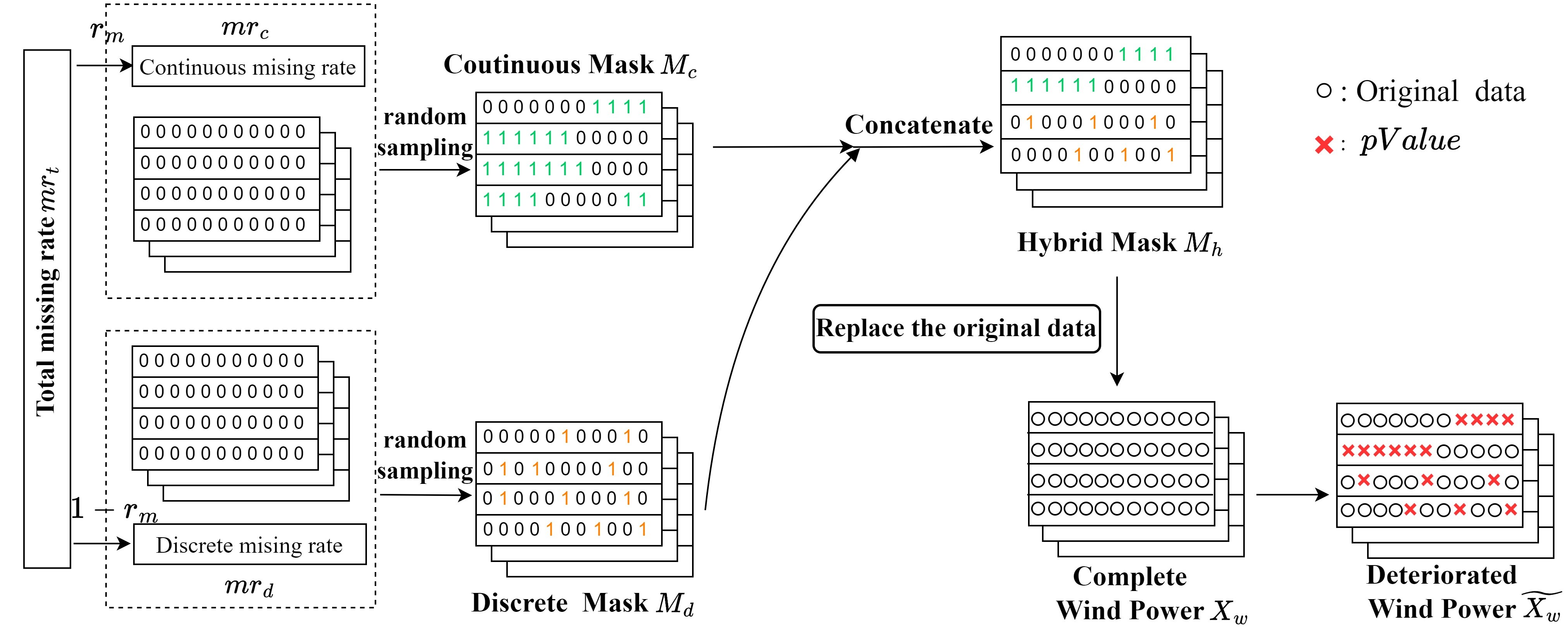}
    \caption{{Construction of missing scenarios using hybrid masking strategy.}}
    \label{fig:enter-label2}
\end{figure}

\begin{figure*}
    \centering
    \includegraphics[width=1\linewidth]{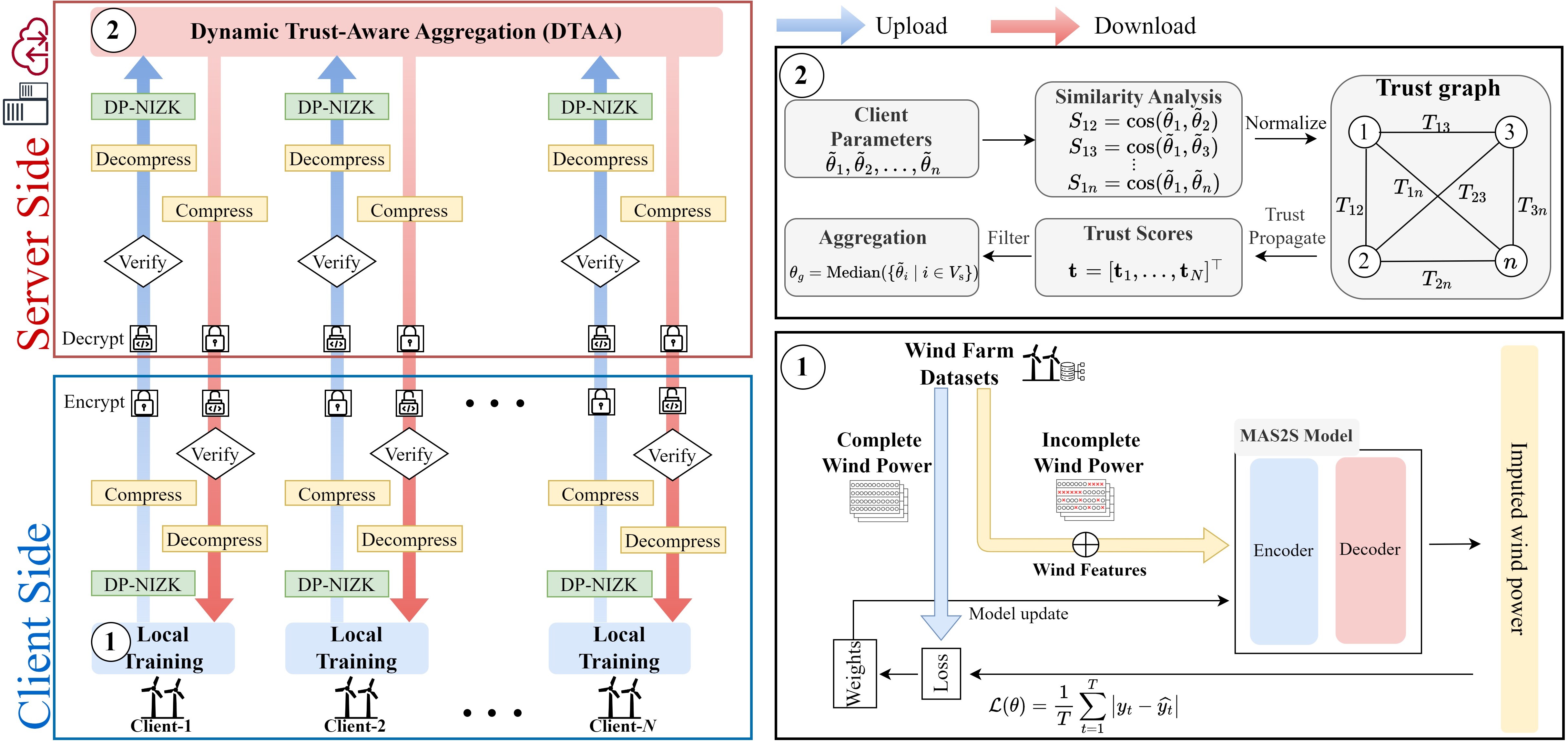}
    \caption{\textcolor{black}{The ZTFed-MAS2S method.}} 
    \label{fig:enter-label3}
    \vspace{-0.4cm}
\end{figure*}

\begin{algorithm}[!ht] 
    \small
    \caption{\textcolor{black}{ZTFed-MAS2S}} %算法标题
    \label{alg:ztfed}
	\begin{algorithmic}[1] %[] 里面的数字为行号 [1]就是一行一个数字
    \REQUIRE Global epoch $T_g$ and its index $t_g$; $I_l$ and $i_l$; Total client ${N}$ and ${n}$; Selected client ${N_e}$ and ${n_e}$; $\bf{n}$; $K$; Client participation rate $E$; local models with parameters $\{\theta_i\}^N_{i=1}$; Global parameters $\theta_g$ on the server.  \\
    \FOR{epoch $t_g=1,2,\ldots, T_g$}
    \STATE Randomly choose $N_e$ clients from all clients $N$ using $E$. 
    \STATE \textbf{Clients execute:}
    \FOR{each selected client ${n_e}$ \textbf{in parallel}}
    \FOR{local training epoch $i_l=1,2,\ldots, I_l$}
    \STATE Update parameters $\theta_{n_e}$ using the MAS2S.
    \ENDFOR
    \ENDFOR
    \IF {$t_g\bmod K = 0$} 
        \STATE Clip parameters $\{\theta_{{n_e}}\}^{N_e}_{n_e=1}$, add noises $\bf{n}$ to the parameters using (16), (17), and generate NIZK proofs $p_f$ according to (20)-(22).
        \STATE Compress and then encrypt the perturbed parameters $\{\widetilde{\theta}_{{n_e}}\}^{N_e}_{n_e=1}$ according to (24)-(26).
        \STATE Upload the $\mathcal{M}_{\text{enc}}^U$ with the $p_f$ to the server.
        \STATE \textbf{Server executes:}
        \STATE Decrypt and decompress the $\mathcal{M}_{\text{enc}}^U$ according to (26).
        \IF {HMAC and NIZK proofs are valid}
            \STATE Aggregate $\{\widetilde{\theta}_{{n_e}}\}^{N_e}_{n_e=1}$ using DTAA and obtain the $\theta_g$ according to (27)-(33).
            \STATE Compress and encrypt the $\theta_g$ according to (24)-(25).
        \ENDIF
        \STATE \textbf{Clients execute:}
        \FOR{each client ${n}$ \textbf{in parallel}}
        \STATE Download the $\mathcal{M}_{\text{enc}}^D$ from the server and decrypt it.
        \IF {HMAC is valid}
        \STATE Decompress and update the local models using $\theta_g$:
        \begin{equation}
        \{\theta_i\}^N_{i=1}\leftarrow\theta_g
        \nonumber
        \end{equation} 
        \ENDIF
        \ENDFOR
    \ENDIF
    \ENDFOR
  \end{algorithmic}
\end{algorithm}

\subsection{ZTFed-MAS2S Architecture}
Based on the theoretical analysis above, the ZTFed-MAS2S method, illustrated in Fig. 3, is developed to impute wind power missing data in wind farms. Further details are provided in Algorithm 2.

\section{Case Studies}
\subsection{Experimental Setup}

The wind energy datasets used for simulations are real-world datasets from the National Renewable Energy Laboratory (NREL)~\cite{draxl2015wind}, comprising 15-minute interval measurements from 16 onshore wind farms in Washington State (2007–2013). It includes wind power, air density, temperature, wind direction, wind speed, and surface air pressure. Each sample has a sequence length of 96, and following~\cite{9158560}, the data are split into 80\% training, 10\% validation, and 10\% testing. All methods are implemented and evaluated on the same hardware platform: an Intel Core i9-10900K CPU, NVIDIA GeForce RTX 3090 GPU, and 32 GB RAM. Experiments are conducted using Python 3.8.18 and TensorFlow 2.6.0.

\textcolor{black}{For fair comparison, baseline methods are evaluated across privacy protection and communication efficiency, aggregation, and imputation. For privacy protection and communication efficiency, we consider standard DP (with $\tau_c$ set to the 95th percentile of parameter norms~\cite{10969624} and $\delta$ set to $1 \times 10^{-4}$), FHE (using the Cheon-Kim-Kim-Song scheme), and TSS (with 3-share splitting). Aggregation mechanisms include FedAvg, MultiKrum (with an adversary ratio of 0.3), and T-Mean (with a trimming rate of 0.1). For imputation, we benchmark statistical (Mean, EM), traditional machine learning (Missforest, kNN), and deep learning (GAIN, Bi-LSTM, Transformer) methods. Detailed imputation configurations are provided in Table~I.}

\textcolor{black}{Unless otherwise specified, all experiments adopt a unified FL configuration: synchronization interval $K=10$, global epochs $T_g=100$, local epochs $I_l=20$, total clients $N=16$, and client participation rate $E=50\%$. Model parameters are compressed with a sparsity ratio of 0.7 and quantized to 4 bits. For DTAA, the number of trusted neighbors is set to $k_g=3$, the tuning coefficient is set to $k_m=3$, the sharpening coefficient is set to $r=2$, and the convergence threshold is set to $\tau_t=10^{-4}$.}

\begin{table}[!t]
    \centering
    \caption{Comparative Analysis Settings}
    \resizebox{\columnwidth}{!}{%
    \scriptsize % 替换 footnotesize 为 scriptsize 进一步减小字体
    \renewcommand{\arraystretch}{0.9} % 减小行间距
    \begin{tabular}{@{\hspace{2pt}}p{2.5cm}p{2.5cm}p{2.5cm}@{\hspace{2pt}}}
    \toprule
    Methods & Parameters & Value/Configuration \\
    \midrule[0.2pt]
    \multirow{5}{*}{MAS2S} 
            & MA\_heads & 2 \\
            & Key\_dim & 32 \\
            & Optimizer & Adam \\
            & Neurons & 128 \\
            & Learning rate & 0.001 \\
    \midrule[0.2pt]
    \multirow{3}{*}{Bi-LSTM} 
            & Neurons & 128 \\
            & Optimizer & Adam \\
            & Learning rate & 0.001 \\
    \midrule[0.2pt]
    \multirow{5}{*}{Transformer} 
            & MA\_heads & 4 \\
            & Key\_dim & 64 \\
            & Ff\_dim & 256 \\
            & Optimizer & Adam \\
            & Learning rate & 0.001 \\
    \midrule[0.2pt]
    \multirow{4}{*}{GAIN} 
            & Optimizer & Adam \\
            & Learning rate & 0.0002 \\
            & Hint rate & 0.99 \\
            & Noise & 0.0-0.1 \\
    \midrule[0.2pt]
    \multirow{2}{*}{kNN} 
            & N\_neighbors & 5 \\
            & Weights\_initializer & Uniform \\
    \midrule[0.2pt]
    \multirow{2}{*}{Missforest} 
            & Max\_iter & 10 \\
            & N\_estimators & 100 \\
    \midrule[0.2pt]
    \multirow{2}{*}{EM} 
            & Loops & 50 \\
            & Tolerance & 10\% \\
    \bottomrule
    \end{tabular}%
    } 
    \vspace{-0.5cm} % 进一步减小下方间距
\end{table}

\subsection{Evaluation Metrics}
To comprehensively evaluate the proposed framework, we adopt the following metrics:

\subsubsection{Imputation Accuracy}We evaluate imputation performance using three standard metrics: mean absolute error (MAE), root mean square error (RMSE), and mean arctangent absolute percentage error (MAAPE). MAAPE mitigates the impact of extreme relative errors, especially when ground truth values are near zero~\cite{malhan2022novel}. It is defined as:
\begin{equation}
   {\rm{MAAPE}}=\frac{1}{n}\sum\limits_{i = 1}^n\arctan\left(\left|\frac{y_i-\widehat{y_i}}{y_i}\right|\right).
\end{equation}

Here, $n$ is the number of missing sample points in the test set, \( \widehat{y_i} \) is the imputed value (only evaluated at missing positions), and \( y_i \) is the corresponding ground truth.

Additionally, RMSE-based sensitivities are introduced to evaluate the methods' robustness \cite{ma2020bi}, detailed as follows:
\begin{equation}
    {S_{mr,rmse}} = \frac{{\sum\limits_{i = 1}^{{n_m}} {\left| {(m{r_i} - \overline {mr} )(rms{e_i} - \overline {rmse} )} \right|} }}{{\sum\limits_{i = 1}^n {{{(m{r_i} - \overline {mr} )}^2}} }},
\end{equation}
\vspace{-0.2cm}
\begin{equation}
{S_{prC,rmse}} = \frac{{\sum\limits_{i = 1}^{{n_c}} {\left| {(pr{C_i} - \overline {prC} )(rms{e_i} - \overline {rmse} )} \right|} }}{{\sum\limits_{i = 1}^{{n_c}} ( pr{C_i} - \overline {prC} {)^2}}},
\end{equation}
where $prC_i$ denotes the proportion of continuous missing in the hybrid missing pattern; ${S_{prC,rmse}}$ is the RMSE-based sensitivity under varying missing rates; ${S_{mr,rmse}}$ is the RMSE-based sensitivity under varying hybrid missing patterns;  $n_m$ and $n_c$ indicate the number of $mr$ and $prC$, respectively. 

\subsubsection{\textcolor{black}{Privacy Protection and Model Utility}}
\textcolor{black}{DP limits the influence of individual samples on model outputs, reducing the risk of membership inference attacks (MIA)~\cite{jiang2021differential}. To evaluate the privacy protection offered by DP, we adopt a black-box MIA threat model~\cite{8844607}, assuming that the attacker has access to the global model $f_{\theta_g}(\cdot)$. The attacker aims to infer the membership status of target samples $\{(X_a, Y_a)\}_{a=1}^A$ by exploiting model predictions.}

\textcolor{black}{For each target sample $(X_a, Y_a)$, the attacker computes the prediction error function $\mathcal{L}_a$ as:}
\begin{equation}
\textcolor{black}{\mathcal{L}_a(X_a, Y_a; \theta_g) = \frac{1}{T} \sum_{t=1}^T \left|f_{\theta_g}(X_{a,t}) - Y_{a,t}\right|,}
\end{equation}
\textcolor{black}{where $(X_{a,t}, Y_{a,t})$ indicates the input-output pair at time step $t$ within sample $a$. The attacker infers the membership status $\hat{m}_a$ of sample $a$ by comparing the error to a threshold $\tau_m$:}
\begin{equation}
\textcolor{black}{\hat{m}_a =
\begin{cases}
1, & \text{if } \mathcal{L}_a(X_a, Y_a; \theta_g) < \tau_m, \\
0, & \text{if } \mathcal{L}_a(X_a, Y_a; \theta_g) \geq \tau_m,
\end{cases}}
\end{equation}
\textcolor{black}{where $\hat{m}_a = 1$ indicates that the attacker classifies the sample as a member, and $\hat{m}_a = 0$ as a non-member. The overall attack performance is evaluated by the MIA success rate (MIA-SR), defined as:}
\begin{equation}
\textcolor{black}{{\rm{MIA-SR}}  = \frac{{\rm{1}}}{{A}}\sum\limits_{{a = 1}}^{A} {\rm{1}} [{{\rm{\hat m}}_{a}}{\rm{ = }}{{\rm{m}}_{a}}] \times 100\% ,}
\end{equation}
\textcolor{black}{where $m_a \in \{0,1\}$ denotes the true membership label of the $a$-th sample, and $A$ is the total number of target samples considered in the evaluation.}

To assess the impact of DP on model performance, a utility metric is defined as:
\begin{equation}
    \text{Utility} = \frac{1 - \text{MAAPE}}{1 - \text{MAAPE(no DP)}} \times 100\%.
\end{equation}
\subsubsection{\textcolor{black}{Communication Overhead}}
\textcolor{black}{In this work, the communication overhead is defined as the total amount of data uploaded and downloaded by clients across all communication rounds, measured in megabytes (MB). Specifically, the communication overhead (CO) in our method is computed as:}
\begin{equation}
\textcolor{black}{{\rm{CO}} = \sum\limits_{r = 1}^R {\left( {\sum\limits_{{n_{e = 1}}}^{{N_e}} {\left( {{\cal M}_{enc,{n_e}}^{U,r} + p_{f,{n_e}}^r} \right) + N \times {\cal M}_{enc}^{D,r}} } \right)}}
\end{equation}
\textcolor{black}{where $R$ is the total number of communication rounds, computed as $R = T_g / K$, with $r$ denoting the round index. $N_e$ and $n_e$ denote the number of clients selected in each communication round and the index of a selected client, respectively. $\mathcal{M}_{\text{enc},n_e}^{U,r}$ is the encrypted upload message from client $n_e$ in round $r$, while $\mathcal{M}_{\text{enc}}^{D,r}$ is the encrypted download message from the server.}

\subsection{\textcolor{black}{Performance Evaluation on Privacy Protection and Communication Efficiency}}

\textcolor{black}{To evaluate the trade-off between DP protection and model performance, we used 500 each of member and non-member samples as MIA attack targets. As shown in Table~II, with the privacy leakage probability $\delta = 1 \times 10^{-4}$ fixed, a smaller $\varepsilon$ leads to a lower MIA success rate, indicating stronger privacy protection. For instance, reducing $\varepsilon$ from 60 to 20 decreases the MIA success rate by 32.58\% (from 87.80 to 59.19). However, this comes at the cost of reduced model utility, which declines by 16.63\% (from 98.79 to 82.36). Without DP (i.e., no noise added), the model achieves the highest utility but faces the highest MIA risk. These results highlight the inherent trade-off in DP: higher $\varepsilon$ improves utility but weakens privacy protection, while lower $\varepsilon$ strengthens privacy but reduces utility. In this task, we set $\varepsilon=40$ as a balanced compromise between privacy protection and model performance~\cite{near2023guidelines}.}

\textcolor{black}{Table~III compares DP-NIZK+CIV and standard DP in terms of RMSE, MIA-Resist (resistance to membership inference attacks), Verifiable (support for verifying correct DP application), and Secure-Trans (protection of model parameters transmission). While both methods achieve comparable RMSE, only DP-NIZK+CIV provides verifiable differential privacy and transmission integrity via non-interactive zero-knowledge proofs and authenticated encryption. Additionally, comparative experiments across client scales ($N \in \{8, 16, 32, 64\}$) were conducted using FHE, TSS, and standard DP as baselines. Fig.~4 shows that the communication overhead of all methods increases as the number of clients grows. Nevertheless, the proposed DP-NIZK+CIV consistently achieves lower communication overhead, reducing costs by at least 61.17\% and 50.13\% compared to FHE and TSS, respectively, across varying client scales. These results demonstrate that DP-NIZK+CIV effectively combines enhanced privacy protection with improved communication efficiency.}

\begin{table}[h]
    \centering
    \caption{\textcolor{black}{Impact of Differential Privacy's Privacy Budget on Model Utility and MIA Success Rate (Mean~(Standard Deviation))}}
    \begin{tabularx}{\linewidth}{@{}X>{\centering\arraybackslash}X>{\centering\arraybackslash}X@{}}
        \toprule
        ($\epsilon,\delta$)-DP         & Utility  & MIA-SR   \\
        \midrule
        No DP                          &100.00(0)       &91.93(0.0181) \\
        (60, $1{\times}10^{-4}$)-DP    &98.79(0.0317) 	&87.80(0.0258) \\
        (40, $1{\times}10^{-4}$)-DP    &97.45(0.0471) 	&80.65(0.0333) \\ 
        (30, $1{\times}10^{-4}$)-DP    &89.31(0.0324) 	&69.80(0.0313) \\ 
        (20, $1{\times}10^{-4}$)-DP    &82.36(0.0299) 	&59.19(0.0548) \\   
        \bottomrule
    \end{tabularx}
    \vspace{-0.4cm}
\end{table}

\begin{table}[h]
    \centering
    \caption{\textcolor{black}{Comparative Evaluation of DP and DP-NIZK+CIV on accuracy and Privacy Protection (Mean~(Standard Deviation))}}
    \resizebox{\columnwidth}{!}{
    \begin{tabular}{@{}lcccccc@{}}
    \toprule
    Method                & RMSE           & MIA-Resist & Verifiable & Secure-Trans \\
    \midrule[0.2pt]
    DP                    & 0.0379(0.0049) & \ding{51}    & \ding{55}   & \ding{55}           \\
    DP-NIZK+CIV  & 0.0384(0.0066) & \ding{51}    & \ding{51}   & \ding{51}           \\
    \bottomrule             
    \end{tabular}}
\end{table}

\begin{figure}[!t]
    \centering
    \includegraphics[width=1\linewidth]{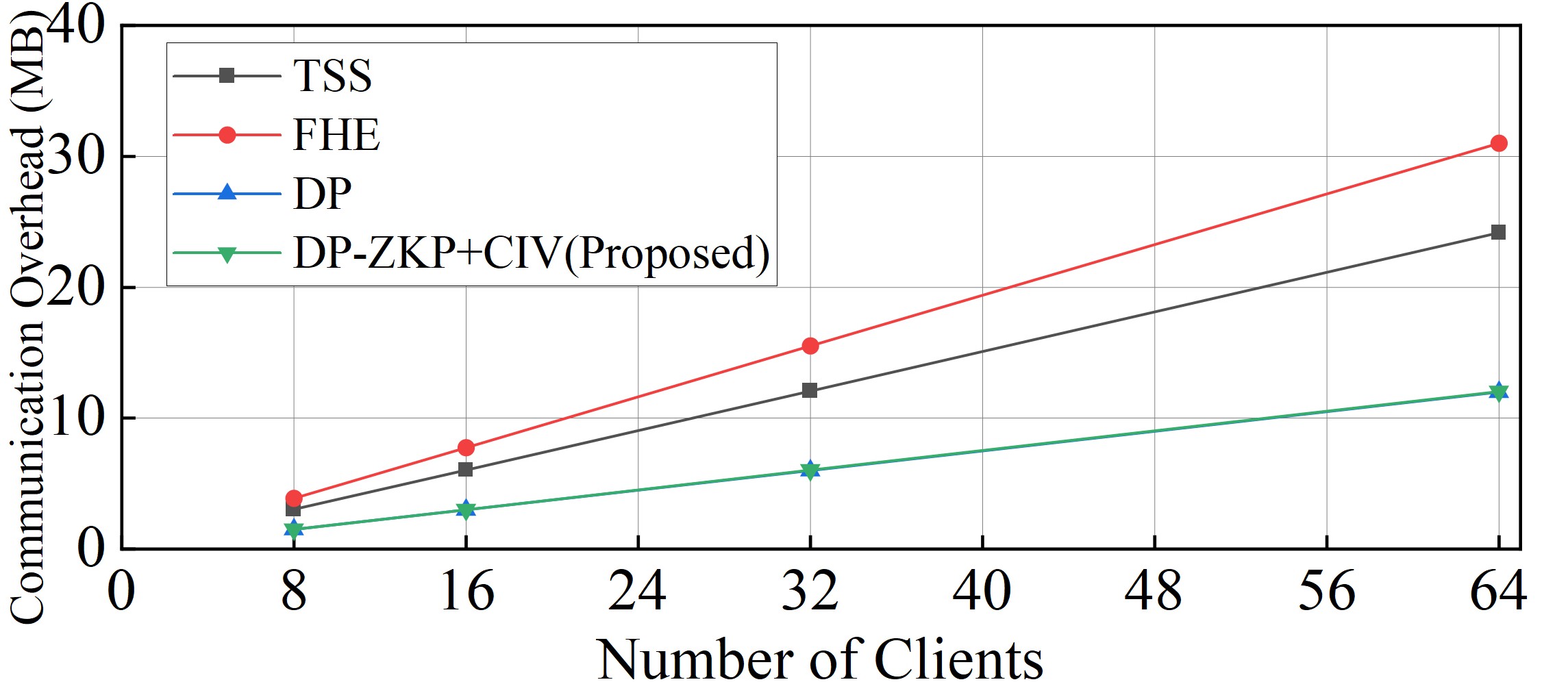}
    \caption{\textcolor{black}{Communication overhead comparison among TSS, FHE, DP, and the proposed DP-NIZK+CIV under different client scales.}}
    \label{fig:enter-label4}
    \vspace{-0.4cm}
\end{figure}

\subsection{\textcolor{black}{Assessment of Aggregation Mechanism}}

\textcolor{black}{To evaluate the robustness of the proposed DTAA method against various client-side anomalies, we use the sign-flipping attack~\cite{10942405} as a representative scenario, where 20\% of each client's model parameters are reversed. DTAA is compared with FedAvg, MultiKrum, and T-Mean under different client scales ($N = \{8, 16, 32, 64\}$) and anomaly rates ($P = \{0, 0.1, 0.3\}$), where $P$ denotes the proportion of anomalous clients among all clients. }

\textcolor{black}{As shown in Fig. 5, DTAA achieves competitive RMSE across most client scales $N$ and anomaly rates $P$, demonstrating superior robustness. In contrast, FedAvg suffers significant degradation as $P$ increases, due to its equal-weight averaging that indiscriminately aggregates abnormal clients. While T-Mean and MultiKrum show some resilience, they have clear limitations: T-Mean applies fixed-ratio trimming, leading to unnecessary discarding of benign model parameters, and MultiKrum requires predefined adversary bounds, both lacking the ability to dynamically adapt to varying anomaly proportions in a ZT environment. As a result, DTAA outperforms T-Mean and MultiKrum, achieving approximately 28.21\% and 17.02\% lower RMSE, respectively, under $N=16$, $P=0.5$.}

\begin{figure}[!t]
    \centering
    \includegraphics[width=1\linewidth]{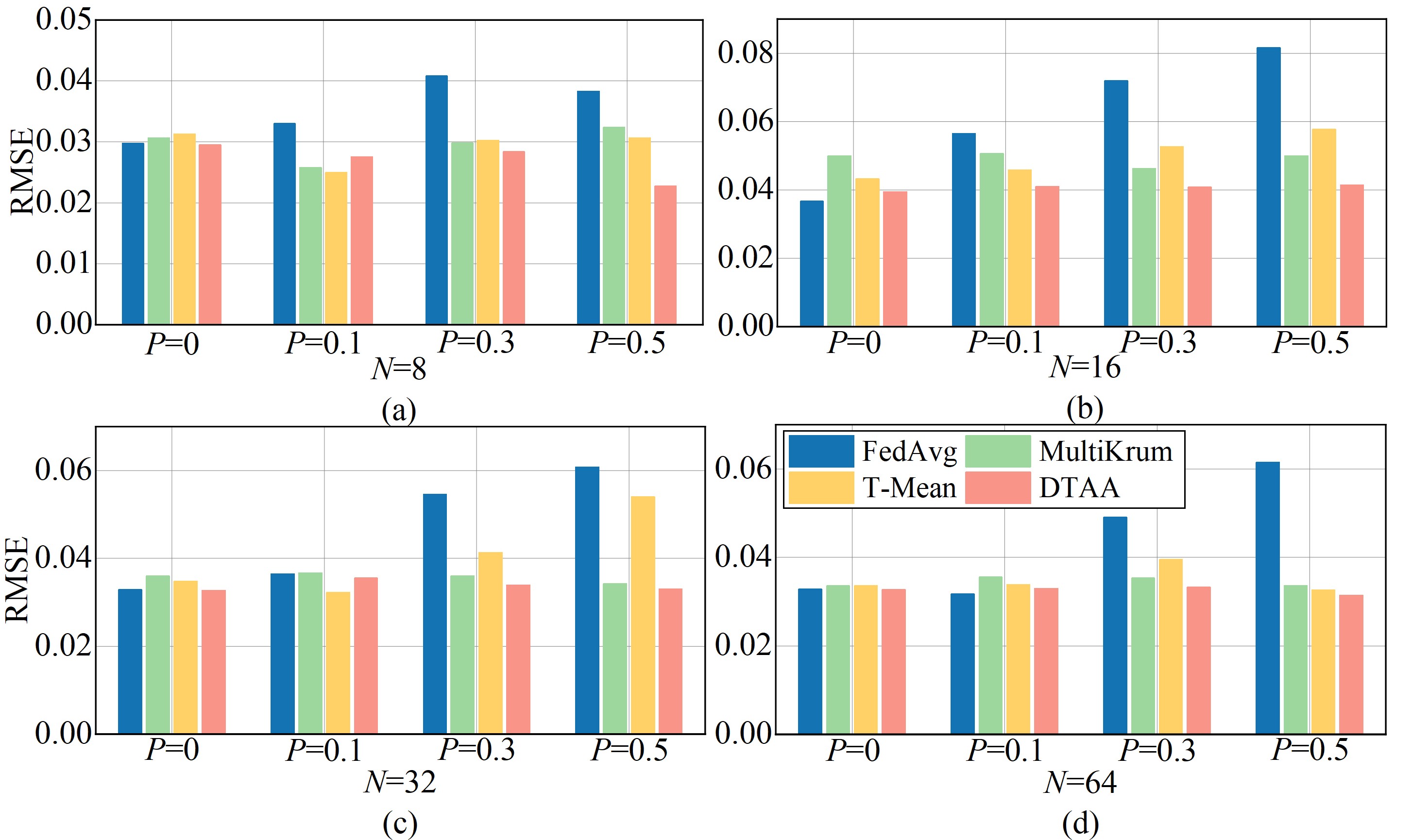}
    \caption{\textcolor{black}{Comparison of aggregation mechanisms under sign-flipping attacks: RMSE across varying anomaly rates ($P$) and client scales ($N$).}}
    \vspace{-0.3cm}
    \label{fig:enter-label5}
\end{figure}

\subsection{\textcolor{black}{Ablation Experiment}}

\textcolor{black}{To investigate the contribution of key components in the proposed method, we conduct ablation experiments by individually removing the multi-head attention mechanism (No MA), the federated learning setting (No Fed), the Dynamic Trust-Aware Aggregation (No DTAA), and the DP-NIZK+CIV privacy-preserving mechanism (No DP-NIZK+CIV). The results are summarized in Table IV.}

\textcolor{black}{Compared to the proposed method, removing the MA leads to the largest RMSE increase (0.040), highlighting its importance in modeling temporal dependencies. The No Fed setting achieves lower errors due to centralized training but sacrifices privacy. Omitting DP-NIZK+CIV slightly improves accuracy, but compromises privacy protection, communication integrity, and verifiability—critical requirements in ZT federated environments.}

% \subsection{Method Robustness under Different FL Parameters}
% To evaluate the robustness of the proposed method under different FL settings, this study simulates various communication scenarios by varying synchronization interval $K$ and client participation rate $E$.

% Fig. 8 illustrates that the RMSE performance of the proposed method initially fluctuates with varying $K$ but gradually stabilizes as the global epochs progress, demonstrating the robustness of ZTFed-MAS2S across different synchronization intervals. Regarding the client participation rate, the curves suggest that while different participation rates do not impact imputation quality, they do affect the convergence speed for a given synchronization interval.
\begin{table}[!t]
    \centering
    
    \caption{\textcolor{black}{Ablation Study of Key Components in the Proposed Method (Mean~(Standard Deviation))}}
    \resizebox{\columnwidth}{!}{
    \begin{tabular}{@{}lcccccc@{}}
    \toprule
    \multirow{1}{*}{Method}                       &MAE  & RMSE  &MAAPE    &$\Delta$RMSE   \\
    \midrule[0.2pt]
    \multirow{1}{*}{Proposed(Baseline)}             &0.0098(0.0015)  &0.0384(0.0049)  &0.1380(0.0238)    &-        \\
    \multirow{1}{*}{No MA}                          &0.0117(0.0010)  &0.0424(0.0073)  &0.1530(0.0124)    &$\uparrow$   0.040       \\
    \multirow{1}{*}{No Fed (Centralization)}        &0.0071(0.0015)  &0.0283(0.0066)  &0.0995(0.0162)    &$\downarrow$ 0.101        \\
    \multirow{1}{*}{No DTAA (FedAvg)}               &0.0108(0.0014)  &0.0403(0.0024)  &0.1469(0.0095)    &$\uparrow$   0.019        \\
    \multirow{1}{*}{No DP-NIZK+CIV}                 &0.0091(0.0009)  &0.0326(0.0041)  &0.1329(0.0144)    &$\downarrow$ 0.058        \\                                                                                       
    \bottomrule             
    \end{tabular}}
    \vspace{-0.4cm}
\end{table}

\subsection{Comparative Experiments with Missing Rates}

Keeping the hybrid missing pattern proportion constant, \textcolor{black}{the missing rate of wind power data in the input samples is set to 20\%, 50\%, 70\%, and 90\%.} By repeatedly testing the methods, the evaluation metrics (MAE, RMSE, MAAPE) on the test sets of all clients are summed and averaged, as shown in Table V.

\textcolor{black}{The evaluation results show that the ZTFed-MAS2S achieves superior overall performance across all missing rates, generally yielding lower MAE, RMSE, and MAAPE compared to other methods.} Under extreme missing rates (90\%), the ZTFed-MAS2S maintains superior performance with only marginal increases in the metrics, showcasing its exceptional resilience to extreme data sparsity. By leveraging its attention mechanism to effectively capture and integrate relevant information from available data points near missing values, our method surpasses the second-best method, \textcolor{black}{with average improvements of 5.71\%, 32.86\%, and 6.19\% in MAE, RMSE, and MAAPE, respectively. At a 90\% missing rate, it outperforms the second-best method by 30.54\%, 42.28\%, and 18.86\% in the same metrics, further demonstrating its superiority. }

\begin{table*}[!t]
    \centering
    \caption{\textcolor{black}{Performance Comparison Across Different Missing Rates under 25\% Discrete Missing Proportion  (Mean~(Standard Deviation))}}
    \resizebox{\linewidth}{!}
    {
    \begin{tabular}{@{}lccccccccc@{}}
    \toprule
    \multirow{2}{*}{Metrics}  & \multirow{2}{*}{Missing}    & \multicolumn{2}{c}{Statistics}& \multicolumn{2}{c}{Machine Learning}& \multicolumn{3}{c}{Deep Learning} &\multirow{2}{*}{\textbf{ZTFed-MAS2S}}\\
        \cmidrule(lr{3pt}){3-4}  \cmidrule(lr{4pt}){5-6} \cmidrule(lr{4pt}){7-9}
     &\centering{rate (\%)}                   &\centering{Mean} &\centering{EM} &\centering{Missforest} &\centering{kNN} &\centering{GAIN} &\centering{Bi-LSTM} &\centering{Transformer}  & \\ 
    \midrule 
    \multirow{5}{*}{MAE}    
                            & 20    &0.2049(0.0011) 	&0.2538(0.0023) 		&0.0966(0.0021) 		&0.0879(0.0003) 		             
                                    &0.0217(0.0005)     &0.0088(0.0004)         &0.0065(0.0005) 		&0.0083(0.0010)  	  
                \\
                            & 50    &0.2251(0.0001) 	&0.2660(0.0001) 		&0.1492(0.0006) 		&0.1575(0.0013) 		             
                                    &0.0360(0.0039)     &0.0295(0.0122)         &0.0077(0.0004)         &0.0091(0.0014) 
                \\
                            & 70    &0.2285(0.0005) 	&0.2663(0.0004) 		&0.1692(0.0011) 		&0.1860(0.0002) 		 
                                    &0.0530(0.0047)     &0.0462(0.0127) 		&0.0112(0.0033) 		&0.0107(0.0016)
               \\
                            & 90    &0.2309(0.0006) 	&0.2632(0.0003) 		&0.1905(0.0015) 		&0.2075(0.0006) 
                                    &0.0751(0.0098) 	&0.0620(0.0133) 	    &0.0167(0.0028) 		&0.0116(0.0023) 	 
              \\
                            & Average   &0.2223 		&0.2623 			    &0.1514 			    &0.1597 			
                                        &0.0464 		&0.0366 		        &0.0105 		        &0.0099	 
              \\
    \midrule[0.1pt] 
    \multirow{5}{*}{RMSE}   
                            & 20    &0.2861(0.0022) 		&0.3771(0.0042) 		&0.1719(0.0030) 		&0.1765(0.0004) 		                        &0.0343(0.0004) 		&0.0411(0.0031) 		&0.0460(0.0014) 		&0.0360(0.0065) 	
                            \\
                            & 50    &0.3176(0.0010) 		&0.3932(0.0007)		    &0.2435(0.0003) 		&0.2699(0.0008) 		 
                                    &0.0496(0.0055) 		&0.0787(0.0303) 		&0.0486(0.0036)		    &0.0348(0.0065) 
                            \\
                            & 70    &0.3263(0.0018) 		&0.3976(0.0004) 		&0.2711(0.0024) 		&0.2980(0.0011) 		 
                                    &0.0689(0.0055) 		&0.1127(0.0255)		    &0.0617(0.0118)		    &0.0408(0.0043)
                            \\
                            & 90    &0.3367(0.0009) 		&0.4005(0.0012) 		&0.3046(0.0037) 		&0.3212(0.0010) 		           
                                    &0.0955(0.0080) 		&0.1375(0.0282) 		&0.0712(0.0230) 		&0.0411(0.0045) 	
                            \\
                            & Average   &0.3167 			&0.3921 			    &0.2478 			    &0.2664 			
                                        &0.0620 			&0.0925 			    &0.0569 			    &0.0382
              \\
    \midrule[0.1pt] 
    \multirow{5}{*}{MAAPE}  
                            & 20    &0.6677(0.0073) 		&0.7397(0.0050) 		&0.4746(0.0022) 		&0.4355(0.0034) 		 
                                    &0.2821(0.0040)		    &0.1628(0.0163) 		&0.1308(0.0380) 		&0.1366(0.0199) 	
                \\     
                            & 50    &0.6767(0.0002) 		&0.7393(0.0030) 		&0.5499(0.0029) 		&0.5427(0.0041) 		 
                                    &0.4051(0.0055) 		&0.2463(0.0408) 		&0.1332(0.0358) 		&0.1347(0.0224) 
                \\              
                            & 70    &0.6776(0.0017) 		&0.7370(0.0003) 		&0.5817(0.0003) 		&0.5934(0.0016) 		 
                                    &0.4813(0.0070) 		&0.3040(0.0431) 		&0.1501(0.0343) 		&0.1398(0.0253) 
               \\  
                            & 90    &0.6788(0.0040) 		&0.7303(0.0028) 		&0.6104(0.0004) 		&0.6339(0.0030) 		 
                                    &0.5573(0.0344)		    &0.3568(0.0360) 		&0.1803(0.0382) 	    &0.1463(0.0285) 
              \\
                            & Average   &0.6752 			&0.7366 			     &0.5542 			    &0.5514 			
                                        &0.4314 			&0.2675 			     &0.1486 			    &0.1394 	
              \\
    
    \bottomrule
    \end{tabular}
    }
    \vspace{-0.2cm}
\end{table*}

\begin{figure}[!t]
    \centering
    \includegraphics[width=1\linewidth]{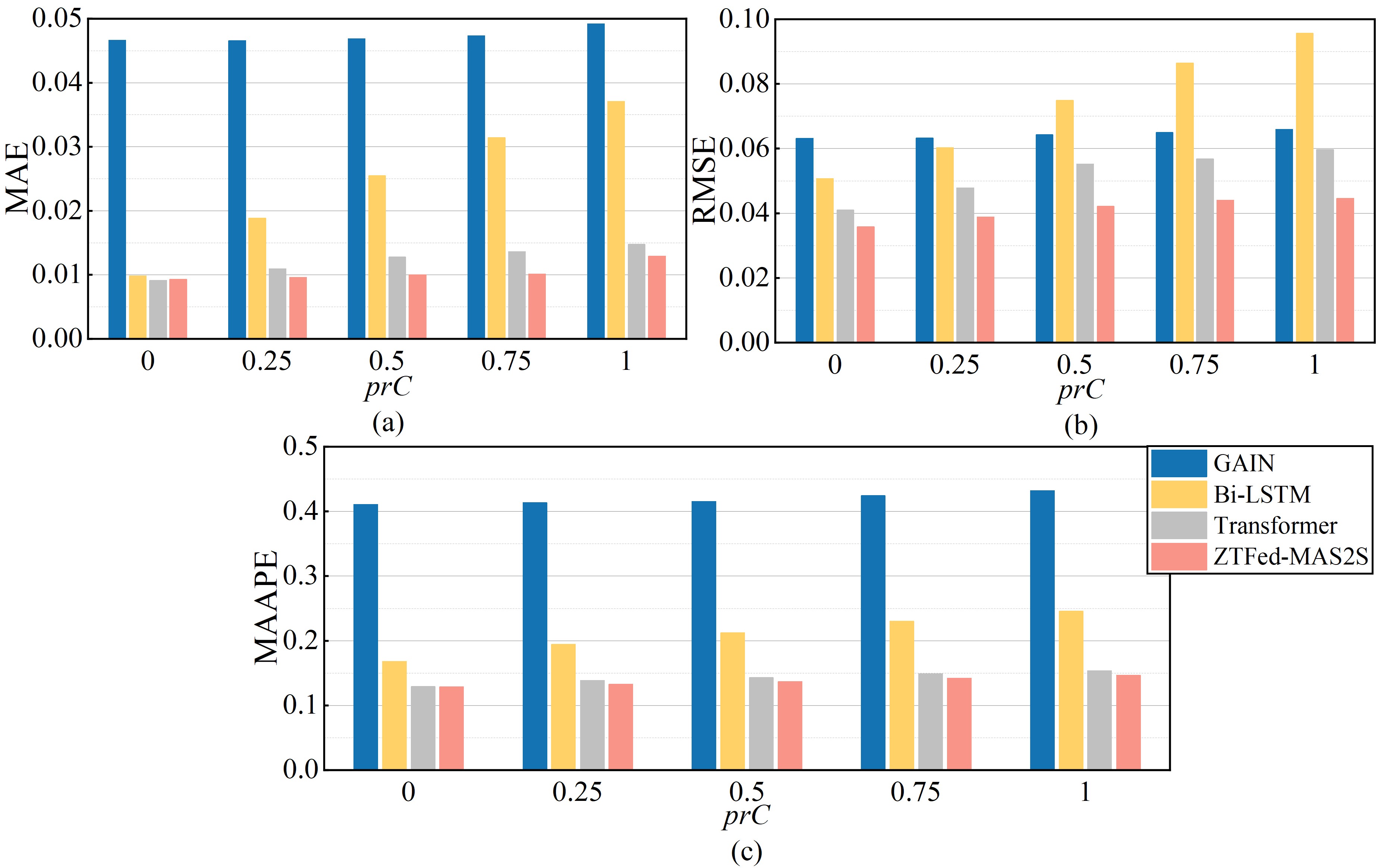}
    \caption{\textcolor{black}{Comparison results with different proportions in the hybrid missing patterns. $prC$ represents the proportion of continuous missing in the hybrid missing pattern.}}
    \label{fig:enter-label6}
    \vspace{-0.4cm}
\end{figure}

\subsection{Comparative Experiments with Hybrid Missing Patterns}

With the environment unchanged, the continuous missing proportion in the hybrid missing patterns decreases from 100\% to 0\%, while the discrete missing proportion rises from 0\% to 100\%, ensuring the missing rate remains constant (by taking the average value), as shown in Fig. 6. The results show that, among the compared deep learning methods:

1) As the continuous missing proportion increases, the performance of all methods declines. The reason for this phenomenon is that a higher proportion of continuous missing data reduces the availability of neighboring data points, making imputation more challenging.

2) Notably, the ZTFed-MAS2S outperforms the other alternatives by leveraging an encoder-decoder architecture and an attention mechanism to effectively capture long-term dependencies. This makes it particularly suitable for handling complex hybrid missing patterns.
\subsection{{Sensitivity Analysis}}

According to (40) and (41), the results of the sensitivity analysis under various missing scenarios are presented in Table VI, where the proposed method is compared against other deep learning-based approaches. \textcolor{black}{Among all the methods, the ZTFed-MAS2S demonstrates the lowest sensitivity values for both metrics, with \( S_{mr,rmse} = 0.0088 \) and \( S_{prC,rmse} = 0.0091 \). Specifically, regarding $S_{mr,rmse}$, our method demonstrates a sensitivity reduction of 76.41\% compared to the second-best performing method, the Transformer.} Similarly, for $S_{prC,rmse}$, our approach achieves a 14.95\% lower sensitivity compared to the second-best method, the GAIN. These results fully demonstrate the superior robustness of the ZTFed-MAS2S in addressing complex missing scenarios.

\begin{table}[!t]
    \centering
    \caption{{\textcolor{black}{Sensitivity Analysis of Different Imputation Methods}}}
    \resizebox{\columnwidth}{!}{
    \begin{tabular}{@{}lccccccc@{}}
    \toprule
    \multirow{1}{*}{Sensitivities}            & GAIN & Bi-LSTM & Transformer & ZTFed-MAS2S    \\
    
    \midrule[0.2pt]                                             
    \multirow{1}{*}{$S_{mr,rmse}$}              &0.0908 	&0.1401 	&0.0373 	&0.0088  \\
    \multirow{1}{*}{$S_{prC, rmse}$}     &0.0107 	&0.0464 	&0.0186 	&0.0091  \\     
    \bottomrule             
    \end{tabular}}
\end{table}

\section{Conclusion}

\textcolor{black}{ZTFed-MAS2S pioneers the integration of verifiable differential privacy and dynamic trust-aware aggregation under a zero-trust architecture for wind data imputation. Comprehensive experiments on the NREL datasets demonstrate the effectiveness and superiority of the proposed method. Specifically:}

\textcolor{black}{1) The DP-NIZK+CIV mechanism ensures verifiable privacy preservation and secure model parameters transmission, reducing communication overhead by at least 61.17\% and 50.13\% compared to the FHE and TSS, respectively. The DTAA enhances robustness against anomalous clients, achieving up to 28.21\% and 17.02\% lower RMSE than the MultiKrum and T-Mean, respectively, under $N=16$, $P=0.5$.}

\textcolor{black}{2) The ZTFed-MAS2S consistently outperforms existing methods in both accuracy and robustness across various missing scenarios. Compared to the Transformer, it achieves 5.71\%, 32.86\%, and 6.19\% higher accuracy in MAE, RMSE, and MAAPE, respectively, with improvements reaching 30.54\%, 42.28\%, and 18.86\% under a 90\% missing rate. Additionally, the method reduces sensitivity to missing rates by 76.41\% compared to the Transformer and to hybrid missing patterns by 14.95\% compared to the GAIN.}

% These results demonstrate the effectiveness and superiority of ZTFed-MAS2S as a secure and robust solution for wind power missing data imputation tasks

While ZTFed provides strong privacy guarantees, its privacy-utility trade-offs require co-optimized designs. Future work will develop adaptive strategies for enhanced robustness against non-independent and identically distributed (non-IID) data heterogeneity in IIoT systems.
\vspace{-0.2cm}

\bibliography{main.bbl}

\begin{thebibliography}{10}
\providecommand{\url}[1]{#1}
\csname url@samestyle\endcsname
\providecommand{\newblock}{\relax}
\providecommand{\bibinfo}[2]{#2}
\providecommand{\BIBentrySTDinterwordspacing}{\spaceskip=0pt\relax}
\providecommand{\BIBentryALTinterwordstretchfactor}{4}
\providecommand{\BIBentryALTinterwordspacing}{\spaceskip=\fontdimen2\font plus
\BIBentryALTinterwordstretchfactor\fontdimen3\font minus \fontdimen4\font\relax}
\providecommand{\BIBforeignlanguage}[2]{{%
\expandafter\ifx\csname l@#1\endcsname\relax
\typeout{** WARNING: IEEEtran.bst: No hyphenation pattern has been}%
\typeout{** loaded for the language `#1'. Using the pattern for}%
\typeout{** the default language instead.}%
\else
\language=\csname l@#1\endcsname
\fi
#2}}
\providecommand{\BIBdecl}{\relax}
\BIBdecl

\bibitem{aaslid2022stochastic}
P.~Aaslid, M.~Korp{\aa}s, M.~M. Belsnes, and O.~B. Fosso, ``Stochastic optimization of microgrid operation with renewable generation and energy storages,'' \emph{IEEE Transactions on Sustainable Energy}, vol.~13, no.~3, pp. 1481--1491, 2022.

\bibitem{10419905}
Z.~Wang and S.~Bu, ``Probabilistic frequency stability analysis considering dynamics of wind power generation with different control strategies,'' \emph{IEEE Transactions on Power Systems}, vol.~39, no.~5, pp. 6412--6425, 2024.

\bibitem{10839636}
W.~Song, J.~Yan, S.~Han, N.~Zang, S.~Liu, C.~Ge, and Y.~Liu, ``A self-supervised pre-learning method for low wind power forecasting,'' \emph{IEEE Transactions on Sustainable Energy}, pp. 1--14, 2025.

\bibitem{10603406}
Q.~Meng, S.~Hussain, F.~Luo, Z.~Wang, and X.~Jin, ``An online reinforcement learning-based energy management strategy for microgrids with centralized control,'' \emph{IEEE Transactions on Industry Applications}, vol.~61, no.~1, pp. 1501--1510, 2025.

\bibitem{9609984}
T.~Wang, H.~Ke, A.~Jolfaei, S.~Wen, M.~S. Haghighi, and S.~Huang, ``Missing value filling based on the collaboration of cloud and edge in artificial intelligence of things,'' \emph{IEEE Transactions on Industrial Informatics}, vol.~18, no.~8, pp. 5394--5402, 2022.

\bibitem{9548664}
X.~Chen, M.~Lei, N.~Saunier, and L.~Sun, ``Low-rank autoregressive tensor completion for spatiotemporal traffic data imputation,'' \emph{IEEE Transactions on Intelligent Transportation Systems}, vol.~23, no.~8, pp. 12\,301--12\,310, 2022.

\bibitem{10106036}
F.~Mouret, A.~Hippert-Ferrer, F.~Pascal, and J.-Y. Tourneret, ``A robust and flexible {EM} algorithm for mixtures of elliptical distributions with missing data,'' \emph{IEEE Transactions on Signal Processing}, vol.~71, pp. 1669--1682, 2023.

\bibitem{10460131}
Y.~Li, L.~Yu, L.~Xing, and F.~Liu, ``Analysis of influencing factors of lane change prediction with data missing,'' \emph{IEEE Transactions on Intelligent Vehicles}, pp. 1--13, 2024.

\bibitem{9518376}
N.~U. Okafor and D.~T. Delaney, ``Missing data imputation on {IoT} sensor networks: Implications for on-site sensor calibration,'' \emph{IEEE Sensors Journal}, vol.~21, no.~20, pp. 22\,833--22\,845, 2021.

\bibitem{liu2022missing}
W.~Liu, C.~Ren, and Y.~Xu, ``Missing-data tolerant hybrid learning method for solar power forecasting,'' \emph{IEEE Transactions on Sustainable Energy}, vol.~13, no.~3, pp. 1843--1852, 2022.

\bibitem{10843972}
X.-Y. Li, Y.~Xu, Q.-X. Zhu, and Y.-L. He, ``Industrial data imputation based on multiscale spatiotemporal information embedding with asymmetrical transformer,'' \emph{IEEE Transactions on Neural Networks and Learning Systems}, pp. 1--12, 2025.

\bibitem{10534859}
J.~V. P~R, N.~Sam~K, V.~T, and A.~C. Kathiresan, ``Development and performance analysis of aquila algorithm optimized {SPV} power imputation and forecasting models,'' \emph{IEEE Transactions on Sustainable Energy}, vol.~15, no.~3, pp. 2103--2114, 2024.

\bibitem{10144489}
L.~Chen, Y.~Xu, Q.-X. Zhu, and Y.-L. He, ``Adaptive multi-head self-attention based supervised {VAE} for industrial soft sensing with missing data,'' \emph{IEEE Transactions on Automation Science and Engineering}, vol.~21, no.~3, pp. 3564--3575, 2024.

\bibitem{10636213}
H.~J. Kim and M.~K. Kim, ``An unsupervised data-mining and generative-based multiple missing data imputation network for energy dataset,'' \emph{IEEE Transactions on Industrial Informatics}, vol.~20, no.~11, pp. 1--12, 2024.

\bibitem{li2022integrated}
Q.~Li, Y.~Xu, B.~S.~H. Chew, H.~Ding, and G.~Zhao, ``An integrated missing-data tolerant model for probabilistic {PV} power generation forecasting,'' \emph{IEEE Transactions on Power Systems}, vol.~37, no.~6, pp. 4447--4459, 2022.

\bibitem{li2023wind}
Y.~Li, R.~Wang, Y.~Li, M.~Zhang, and C.~Long, ``Wind power forecasting considering data privacy protection: A federated deep reinforcement learning approach,'' \emph{Applied Energy}, vol. 329, p. 120291, 2023.

\bibitem{10579800}
R.~Fotohi, F.~Shams~Aliee, and B.~Farahani, ``A lightweight and secure deep learning model for privacy-preserving federated learning in intelligent enterprises,'' \emph{IEEE Internet of Things Journal}, vol.~11, no.~19, pp. 31\,988--31\,998, 2024.

\bibitem{li2022detection}
Y.~Li, X.~Wei, Y.~Li, Z.~Dong, and M.~Shahidehpour, ``Detection of false data injection attacks in smart grid: A secure federated deep learning approach,'' \emph{IEEE Transactions on Smart Grid}, vol.~13, no.~6, pp. 4862--4872, 2022.

\bibitem{fotohi2024decentralized}
R.~Fotohi, F.~S. Aliee, and B.~Farahani, ``Decentralized and robust privacy-preserving model using blockchain-enabled federated deep learning in intelligent enterprises,'' \emph{Applied Soft Computing}, vol. 161, p. 111764, 2024.

\bibitem{10121613}
K.~Wei, J.~Li, M.~Ding, C.~Ma, Y.-S. Jeon, and H.~V. Poor, ``Covert model poisoning against federated learning: Algorithm design and optimization,'' \emph{IEEE Transactions on Dependable and Secure Computing}, vol.~21, no.~3, pp. 1196--1209, 2024.

\bibitem{wang2025federated}
T.~Wang, Z.~Zheng, and F.~Lin, ``Federated learning framework based on trimmed mean aggregation rules,'' \emph{Expert Systems with Applications}, p. 126354, 2025.

\bibitem{9721118}
K.~Pillutla, S.~M. Kakade, and Z.~Harchaoui, ``Robust aggregation for federated learning,'' \emph{IEEE Transactions on Signal Processing}, vol.~70, pp. 1142--1154, 2022.

\bibitem{hijazi2023secure}
N.~M. Hijazi, M.~Aloqaily, M.~Guizani, B.~Ouni, and F.~Karray, ``Secure federated learning with fully homomorphic encryption for {IoT} communications,'' \emph{IEEE Internet of Things Journal}, vol.~11, no.~3, pp. 4289--4300, 2023.

\bibitem{8765347}
G.~Xu, H.~Li, S.~Liu, K.~Yang, and X.~Lin, ``Verifynet: Secure and verifiable federated learning,'' \emph{IEEE Transactions on Information Forensics and Security}, vol.~15, pp. 911--926, 2020.

\bibitem{stafford2020zero}
V.~Stafford, ``Zero trust architecture,'' \emph{NIST special publication}, vol. 800, no. 207, pp. 800--207, 2020.

\bibitem{bahdanau2014neural}
D.~Bahdanau, ``Neural machine translation by jointly learning to align and translate,'' \emph{in Proc. 3rd Int. Conf. Learn. Representa}, 2015.

\bibitem{10111057}
O.~Rubasinghe, X.~Zhang, T.~K. Chau, Y.~H. Chow, T.~Fernando, and H.~H.-C. Iu, ``A novel sequence to sequence data modelling based {CNN-LSTM} algorithm for three years ahead monthly peak load forecasting,'' \emph{IEEE Transactions on Power Systems}, vol.~39, no.~1, pp. 1932--1947, 2024.

\bibitem{10966041}
Z.~Xing, Z.~Zhang, Z.~Zhang, Z.~Li, M.~Li, J.~Liu, Z.~Zhang, Y.~Zhao, Q.~Sun, L.~Zhu, and G.~Russello, ``Zero-knowledge proof-based verifiable decentralized machine learning in communication network: A comprehensive survey,'' \emph{IEEE Communications Surveys \& Tutorials}, pp. 1--1, 2025.

\bibitem{9069945}
K.~Wei, J.~Li, M.~Ding, C.~Ma, H.~H. Yang, F.~Farokhi, S.~Jin, T.~Q.~S. Quek, and H.~Vincent~Poor, ``Federated learning with differential privacy: Algorithms and performance analysis,'' \emph{IEEE Transactions on Information Forensics and Security}, vol.~15, pp. 3454--3469, 2020.

\bibitem{draxl2015wind}
C.~Draxl, A.~Clifton, B.-M. Hodge, and J.~McCaa, ``The wind integration national dataset (wind) toolkit,'' \emph{Applied Energy}, vol. 151, pp. 355--366, 2015.

\bibitem{9158560}
S.~Yang, M.~Dong, Y.~Wang, and C.~Xu, ``Adversarial recurrent time series imputation,'' \emph{IEEE Transactions on Neural Networks and Learning Systems}, vol.~34, no.~4, pp. 1639--1650, 2023.

\bibitem{10969624}
C.~Wei, W.~Li, C.~Gong, and W.~Chen, ``{DC-SGD}: Differentially private {SGD} with dynamic clipping through gradient norm distribution estimation,'' \emph{IEEE Transactions on Information Forensics and Security}, vol.~20, pp. 4498--4511, 2025.

\bibitem{malhan2022novel}
P.~Malhan and M.~Mittal, ``A novel ensemble model for long-term forecasting of wind and hydro power generation,'' \emph{Energy Conversion and Management}, vol. 251, p. 114983, 2022.

\bibitem{ma2020bi}
J.~Ma, J.~C. Cheng, F.~Jiang, W.~Chen, M.~Wang, and C.~Zhai, ``A bi-directional missing data imputation scheme based on {LSTM} and transfer learning for building energy data,'' \emph{Energy and Buildings}, vol. 216, p. 109941, 2020.

\bibitem{jiang2021differential}
B.~Jiang, J.~Li, G.~Yue, and H.~Song, ``Differential privacy for industrial internet of things: Opportunities, applications, and challenges,'' \emph{IEEE Internet of Things Journal}, vol.~8, no.~13, pp. 10\,430--10\,451, 2021.

\bibitem{8844607}
L.~Song, R.~Shokri, and P.~Mittal, ``Membership inference attacks against adversarially robust deep learning models,'' in \emph{2019 IEEE Security and Privacy Workshops (SPW)}, 2019, pp. 50--56.

\bibitem{near2023guidelines}
J.~P. Near, D.~Darais, N.~Lefkovitz, G.~Howarth \emph{et~al.}, ``Guidelines for evaluating differential privacy guarantees,'' \emph{National Institute of Standards and Technology, Tech. Rep}, pp. 800--226, 2023.

\bibitem{10942405}
L.~Yang, Y.~Miao, Z.~Liu, Z.~Liu, X.~Li, D.~Kuang, H.~Li, and R.~H. Deng, ``Enhanced model poisoning attack and multi-strategy defense in federated learning,'' \emph{IEEE Transactions on Information Forensics and Security}, vol.~20, pp. 3877--3892, 2025.

\end{thebibliography}
\bibliographystyle{IEEEtran}

\end{document}